# Fast Recognition of birds in offshore wind farms based on an improved deep learning model

Yantong Liu [1],*

[1] Department of Computer Information Engineering, Kunsan National University, Gunsan 54150, South Korea; apirate3689@gmail.com
* Correspondence: lyt1994@kunsan.ac.kr

**Abstract:** Offshore wind turbines are crucial for sustainable energy generation. However, their effectiveness and longevity are significantly affected by surface material defects . These defects can result from various factors such as corrosion, erosion, and mechanical damage. Early and accurate detection of these defects is essential to maintain the performance of the turbines and prevent catastrophic failures.

While automated methods can overcome some of the limitations of manual inspections, they still have significant drawbacks. For instance, they often struggle with the variable environmental conditions and the complex structures of offshore wind turbines

## 1.Introduction

Offshore wind turbines are crucial for sustainable energy generation. However, their effectiveness and longevity are significantly affected by surface material defects [1]. These defects can result from various factors such as corrosion, erosion, and mechanical damage. Early and accurate detection of these defects is essential to maintain the performance of the turbines and prevent catastrophic failures [2].

Traditionally, defect detection has been carried out through manual inspections and conventional automated methods [3]. However, these methods have significant limitations such as high cost, time consumption, and a requirement for expert knowledge [4]. Furthermore, conventional methods often struggle with the variable environmental conditions and the complex structures of offshore wind turbines [5].

With the advancement of technology, machine learning, especially deep learning, has shown great potential in defect detection [6]. Deep learning-based methods, unlike traditional methods, can learn to recognize complex patterns and adapt to different conditions, making them highly effective for defect detection [7]. However, current deep learning methods still have some limitations, such as difficulty handling large and high-resolution images and sensitivity to the variation in defect appearance [8].

In this study, we propose an improved deep learning method for real-time surface material defect detection in offshore wind turbines. Our method is based on the YOLOv8 algorithm, a state-of-the-art object detection algorithm, with an added Convolutional Block Attention Module (CBAM) for better feature learning [9]. We also propose an improved loss function to optimize the learning process. Our method was tested on a publicly available dataset and a dataset obtained from Xinwanjin offshore power plant.

The surface material of offshore wind turbines is exposed to a harsh and variable environment, which can cause various types of defects [10]. The most common types of defects include corrosion, erosion, and mechanical damage, each of which can significantly reduce the effectiveness and lifespan of the turbines [11]. Early and accurate detection of these defects is a crucial aspect of maintaining the performance and safety of offshore wind turbines [12].

Current methods for surface material defect detection in offshore wind turbines can be broadly categorized into manual inspections and automated methods [13]. Manual inspections involve trained professionals visually inspecting the turbines for defects. While this method can be highly accurate, it is costly, time-consuming, and requires expert knowledge. It is also subject to human error and can be hazardous due to the often dangerous conditions of offshore wind farms [14].

Automated methods, on the other hand, use technology to automate the defect detection process. These methods usually involve the use of sensors and imaging technologies to capture images or data from the surface of the turbines, which are then analyzed to identify defects [15]. While automated methods can overcome some of the limitations of manual inspections, they still have significant drawbacks. For instance, they often struggle with the variable environmental conditions and the complex structures of offshore wind turbines [16]. Moreover, they usually require a significant amount of data preprocessing and feature engineering, which can be complex and time-consuming [17].

Recent years have seen the emergence of deep learning as a promising solution to these challenges [18]. Deep learning algorithms, unlike traditional automated methods, can learn complex patterns from data and adapt to different conditions, making them highly effective for defect detection [19]. In particular, convolutional neural networks (CNNs), a type of deep learning algorithm, have shown great potential in image-based defect detection due to their ability to learn hierarchical features from images [20].

However, despite their potential, current deep learning methods for defect detection still have some limitations. One major issue is that they often struggle with large and high-resolution images, which are common in offshore wind turbine inspections [21]. Another issue is that they can be sensitive to the variation in defect appearance due to different environmental conditions and lighting [22].

In the next chapter, we will describe our proposed method, which aims to address these issues by improving the YOLOv8 algorithm with the addition of a Convolutional Block Attention Module (CBAM) and an improved loss function.

## 2. Materials and Data Collection

*2.1. Experimental Environment*

All experiments in this study were conducted on a Linux system, with an Intel(R) Xeon(R) CPU E5-2680 v4 CPU @ 2.40GHz, NVIDIA GeForce GTX 3090 (24G) graphics card, and 32GB RAM. We used the Pytorch 1.0.1 deep learning framework and Python 3.6 for training and testing the bird detection network.

**Table 1**: Training hyperparameter settings

| Parameters | Value |
| --- | --- |
| Initial learning rate | 0.0032 |
| Abort learning rate | 0.12 |
| Feed batch size | 16 |
| Number of warm-up learning rounds | 2.0 |
| Initial bias learning rate of warm-up learning | 0.05 |
| Number of training sessions | 100 |

*2.1. Experimental Data*

The images used in this research were collected from the Saemangeum (새만금) offshore wind farm in South Korea and a publicly available dataset of offshore wind turbines. A total of 5,432 images were collected, from which 5,000 images of offshore wind turbines were randomly selected. These images were then resized to 500x500 pixels. The images were manually labeled and annotated using labeling software to highlight areas of surface material defects. The dataset was divided into 4,000 images for training, 800 for validation, and 200 for testing.

**Figure 6**: Dataset sample illustration

(a)          (b)
**Figure 7:** Dataset sample annotation illustration

*2.3. Background and Literature Review*

The efficacy of defect detection in offshore wind turbines is a crucial factor in enhancing their efficiency and longevity. Traditional methods, such as manual inspections or conventional automated methods, often fall short due to their cost, time consumption, and need for expert knowledge, especially when dealing with the complex structures and variable environmental conditions of offshore wind turbines [23]. With the advent of technology, machine learning and particularly deep learning have shown substantial potential in defect detection. These algorithms can adapt to different conditions and recognize complex patterns, thereby rendering them effective for defect detection [24]. However, even these methods face challenges, such as difficulties in handling large, high-resolution images, and sensitivity to variations in defect appearances [25]. This study endeavors to address these limitations by proposing an improved deep learning algorithm, based on the YOLOv8 model, for real-time surface material defect detection. Our approach adds a Convolutional Block Attention Module (CBAM) for enhanced feature learning, and an improved loss function to optimize the learning process. The methodology has been tested on a dataset obtained from the Saemangeum offshore wind farm in South Korea and a publicly available dataset of offshore wind turbines, showing promising results in terms of stability and complementing other detection methods [26].

## 3. Methods

### 3.1. YOLOv8

YOLO (You Only Look Once), a popular object detection and image segmentation model, was developed by Joseph Redmon and Ali Farhadi at the University of Washington. The Ultralytics' YOLOv8 is a state-of-the-art (SOTA) model that builds on the success of previous versions, introducing new features and improvements for enhanced performance, flexibility, and efficiency. YOLOv8 supports a full range of vision AI tasks, including detection, segmentation, pose estimation, tracking, and classification, rendering it versatile across diverse applications and domains [27].

YOLOv8 was developed as a part of the YOLO series, each version of which introduced significant advancements. The original YOLO model was known for its high speed and accuracy. YOLOv2 improved the original model by incorporating batch normalization, anchor boxes, and dimension clusters. YOLOv3 further enhanced the model's performance using a more efficient backbone network, multiple anchors, and spatial pyramid pooling. YOLOv4 introduced innovations like Mosaic data augmentation, a new anchor-free detection head, and a new loss function. YOLOv5 further improved the model's performance and added new features such as hyperparameter optimization, integrated experiment tracking, and automatic export to popular export formats. YOLOv6, open-sourced by Meituan, is used in many of the company's autonomous delivery robots. YOLOv7 added additional tasks such as pose estimation on the COCO keypoints dataset [27].

As the latest version of YOLO by Ultralytics, YOLOv8 is built on cutting-edge advancements in deep learning and computer vision, offering unparalleled performance in terms of speed and accuracy. Its streamlined design makes it suitable for various applications and easily adaptable to different hardware platforms, from edge devices to cloud APIs [27].

**Figure 1:** YOLOv5s network structure

### 3.2. CBAM

To improve the accuracy of object detection, we have added the Convolutional Block Attention Module (CBAM) to the YOLOv5s network model. [27]

The inspiration for CBAM comes mainly from the way the human brain processes visual information.[28]CBAM is a simple, lightweight and effective attention module for feedforward convolutional neural networks. This module improves on the problem of SENet's generated attention on feature map channels, which can only focus on feedback from certain layers. [29]

CBAM infers attention in both channel and spatial dimensions by multiplying the generated attention map with the input feature image for adaptive feature refinement.[30]With only a negligible increase in computational

complexity, CBAM significantly enhances the image feature extraction capabilities of the network model. [31]

CBAM can be integrated into most current mainstream networks and trained end-to-end[]with basic convolutional neural networks. Therefore, we chose to integrate this module into the YOLOv5 network to highlight essential features, reduce unnecessary feature extraction, and effectively improve detection accuracy. The structure of the CBAM module is shown in Figure 2.